\documentclass{article}

\PassOptionsToPackage{numbers, compress}{natbib}

\usepackage[preprint]{neurips_2019}

\usepackage[utf8]{inputenc} 
\usepackage[T1]{fontenc}    
\usepackage{hyperref}       
\usepackage{url}            
\usepackage{booktabs}       
\usepackage{amsfonts}       
\usepackage{nicefrac}       
\usepackage{microtype}      
\usepackage{amsmath}
\usepackage{amssymb}
\usepackage{color}
\usepackage{graphicx}
\usepackage{float}
\usepackage{subfigure}
\usepackage{microtype}
\usepackage{multirow,adjustbox}
\usepackage{caption}
\newcommand{\x}{{\rm\bf x}}      
\newcommand{\f}{{\rm\bf f}}      
\newcommand{\loss}{{\mathcal L}} 
\newcommand{\ie} {\emph{i.e. }}
\newcommand{\eg} {\emph{e.g. }}
\newcommand{\vs} {\emph{v.s. }}
\newcommand{\wrt} {\emph{w.r.t. }}

\newcommand{\wo} {\emph{w/o }}

\title{Residual Knowledge Distillation}

\author{
  Mengya Gao\textsuperscript{1},
  Yujun Shen\textsuperscript{2},
  Quanquan Li\textsuperscript{3},
  Chen Change Loy\textsuperscript{4} \\
  \textsuperscript{1}Tianjin University \quad
  \textsuperscript{2}The Chinese University of Hong Kong \\
  \textsuperscript{3}SenseTime Group Limited \quad
  \textsuperscript{4}Nanyang Technological University \\
  {\tt\small
    daisy@tju.edu.cn,
    sy116@ie.cuhk.edu.hk,
    liquanquan@sensetime.com,
    ccloy@ntu.edu.sg
  }
}

\begin{document}

\maketitle

\begin{abstract}
Knowledge distillation (KD) is one of the most potent ways for model compression.
The key idea is to transfer the knowledge from a deep teacher model ($T$) to a shallower student ($S$).
However, existing methods suffer from performance degradation due to the substantial gap between the learning capacities of $S$ and $T$. 
To remedy this problem, this work proposes Residual Knowledge Distillation (RKD), which further distills the knowledge by introducing an \emph{assistant} ($A$).
Specifically, $S$ is trained to mimic the feature maps of $T$, and $A$ aids this process by learning the \emph{residual error} between them.
In this way, $S$ and $A$ complement with each other to get better knowledge from $T$.
Furthermore, we devise an effective method to derive $S$ and $A$ from a given model \emph{without} increasing the total computational cost.
Extensive experiments show that our approach achieves appealing results on popular classification datasets, CIFAR-100 and ImageNet, surpassing state-of-the-art methods.
\end{abstract}

\section{Introduction}\label{sec:introduction}
%
%
%
Knowledge distillation (KD) \cite{bucilua2006model,hinton2014distilling} is a popular solution to model compression.
%
In general, KD starts with training a large model, called teacher ($T$), to achieve appealing performance, and then employs a lower-capacity one, termed as student ($S$), to learn knowledge from $T$.
In this way, $S$ is supposed to produce similar prediction as $T$ but with faster speed and less memory consumption.
%

\begin{figure}[t]
  \centering
  \includegraphics[width=1.0\linewidth]{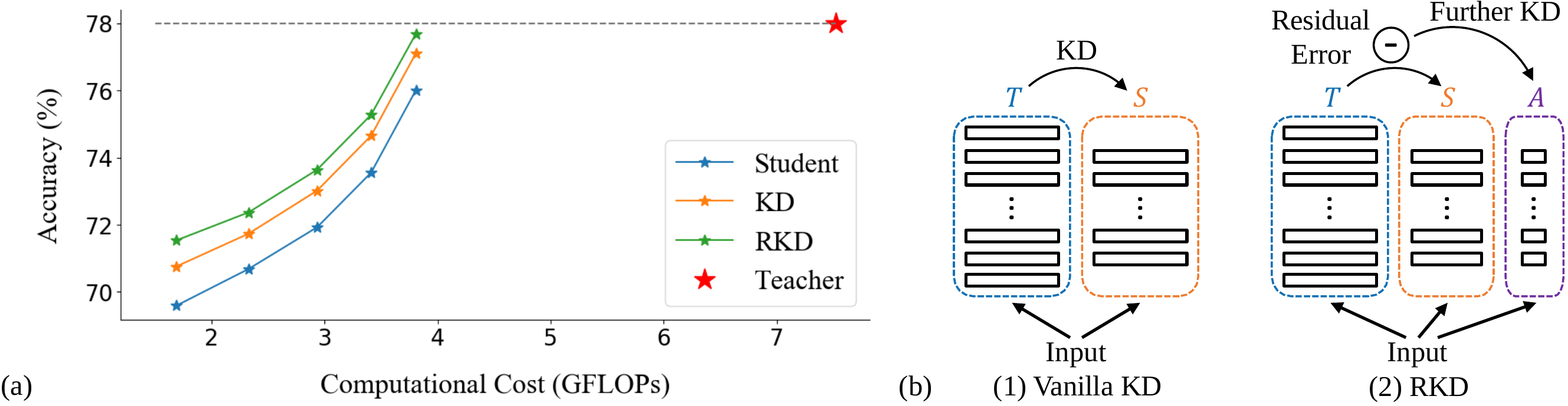}
  \captionsetup{font=small}
  \caption{
    (a) represents the relationships between model capacity and performance on ImageNet across different methods.
    Here, model capacity is measured by the computational cost in unit of Giga floating-point operations (GFLOPs).
    Red star on the top-right corner indicates the teacher model.
    (b) shows the comparison between vanilla knowledge distillation (KD) method with our proposed RKD.
    An assistant model is \emph{separated from} the student model to learn the residual error between the feature maps of student and teacher.
    The total computational cost of $S$ and $A$ in RKD is ensured to \emph{not} increase compared to the single $S$ in vanilla KD.
    For each model, the number and width of rectangles indicate the computing resources required.
    Better viewed in color.
  }
  \label{fig:teaser}
  \vspace{-10pt}
\end{figure}

To achieve more effective knowledge transfer from $T$ to $S$, many attempts have been made \cite{romero2015fitnets,huang2017like,yim2017gift,zagoruyko2017paying}.
However, this problem is far from being solved.
As shown in Fig.\ref{fig:teaser} (a), even though KD (orange line) helps improve the performance of student (blue line), there still exists a huge gap compared to the teacher (dashed grey line).
This is mainly caused by two reasons.
First, $S$ has a much weaker representation ability than $T$.
As can be observed in Fig.\ref{fig:teaser} (a), although the performance of $S$ can gradually approximate that of $T$ by increasing its capacity, the inherent discrepancy between the model sizes of $S$ and $T$ still prevents the student to fully acquire knowledge from teacher.
%
%
%
Second, there lacks of an effective strategy to distill the knowledge inside $T$.
Previous KD methods typically use one teacher to supervise one student, resulting in a one-to-one learning.
That is to say, the distillation process happens only once when $S$ is optimized to mimic $T$, as shown in Fig.\ref{fig:teaser} (b.1).
Recall that the capacity of $S$ is far behind that of $T$.
Therefore, such one-time transfer scheme may lead to information lost to some extent.

To ease the process of knowledge transfer, we present a simple yet novel method, called Residual Knowledge Distillation (RKD), whose concept diagram is depicted in Fig.\ref{fig:teaser} (b.2).
Specifically, feature maps are treated as knowledge in this work, and $S$ is trained with the goal of producing identical feature maps as in $T$.
Instead of hoping the student to achieve the ideal mimicking  on its own, an assistant model $A$ is derived from it to help learn the residual error between the feature maps of $S$ and $T$.
It is worth noting that the total computational cost will \emph{not} increase.
With such transferring scheme, the final feature map by summing up the outputs of $S$ and $A$ is more indistinguishable from $T$, hence narrowing down the performance gap between $S$ and $T$, as shown in Fig.\ref{fig:teaser} (a).
%

The \textbf{contributions} of this work are summarized as follows:
\begin{itemize}
  \vspace{-6pt}
  \setlength{\itemsep}{4pt}
  \setlength{\parsep}{0pt}
  \setlength{\parskip}{0pt}
  \item
  We propose a novel model compression method, RKD, by introducing the residual learning strategy into the conventional KD method.
  Different from previous methods that perform a single round of knowledge distillation, RKD distills the knowledge for a second time by training an assistant model $A$ to learn the residual error between $S$ and $T$.
  In addition, this idea can also be applied to other existing KD methods, suggesting its generalization ability.
  \item
  We propose a simple yet effective model separation technique to derive $A$ and $S$ from the same model without increasing the total capacity.
  We further empirically study the way to allocate the computational resources between $A$ and $S$, and find that $A$ only requires a lightweight model (around 1/10 model size of $S$) to mimic the residual error perfectly.
  \item
  We evaluate the proposed RKD on two commonly used classification datasets, CIFAR-100 \cite{krizhevsky2009learning} and ImageNet \cite{deng2009imagenet}, and achieve superior performance compared to the state-of-the-art approaches.
  For example, in the experiments of transferring knowledge from ResNet-34 \cite{he2016deep} to ResNet-18, RKD achieves 2.0\% top-1 accuracy improvement on ImageNet dataset, significantly outperforming the second competitor with 1.2\% improvement.
\end{itemize}

\section{Related Work}\label{sec:related-work}


\textbf{Knowledge Type in KD.}
Existing methods have designed various types of knowledge to improve KD.
\citet{ba2014deep} treated the hard label predicted by $T$ as the underlying knowledge, with the assumption that the well-trained $T$ has already eliminated some label errors contained in the ground-truth data.
\citet{hinton2014distilling} argued that the soft label produced by $T$, \ie the classification probabilities, can provide richer information.
Some work extracted the knowledge from $T$ by processing the hidden feature map.
\citet{zagoruyko2017paying} averaged the feature map across channel dimension to obtain spatial attention map, \citet{yim2017gift} defined inter-layer flow by computing the inner product of two feature maps, and \citet{lee2018self} improved this idea with singular value decomposition (SVD).
A recent work \cite{gao2018embarrassingly} demonstrated the effectiveness of mimicking feature map directly in KD task.
Similarly, this work also applies feature map as the knowledge, since the residual error between feature maps is well defined.
However, our proposed RKD is not limited to learning feature map, but can also work together with other types of knowledge, \eg attention map, as long as there is a way to compute the residual error.

\textbf{Transferring Strategy in KD.}
Besides knowledge type, transferring strategy is another widely studied direction in KD.
\citet{romero2015fitnets} presented FitNets which selects a hidden layer from $T$ and $S$ respectively to be hint layer and guided layer.
Through pre-training the guided layer with the hint layer as supervision, $S$ is able to get a better initialization than trained from scratch.
Net2net \cite{chen2016net2net} also explored the way to initialize the parameters of $S$ by proposing function-preserving transformation, which makes it possible to directly reuse an already trained model.
In addition to initialization, there are some methods combining KD with other techniques to transfer knowledge from $T$ to $S$ more efficiently.
\citet{Belagiannis2018Adversarial} involved adversarial learning into KD by employing a discriminator to tell whether the outputs of $S$ and $T$ are close enough, \citet{Ashok2018N2N} exploited reinforcement learning to find out the best network structure of $S$ under the guidance of $T$, and \citet{wang2018progressive,gao2018embarrassingly} referred to the idea of progressive learning to make knowledge transferred step by step.
Nevertheless, all of the above methods use a single model, $S$, to learn from $T$, and the knowledge is distilled only once.
Considering the difference between the learning capacities of $S$ and $T$, there is no guarantee that $S$ can obtain enough information to reproduce the performance of $T$ via such one-time transfer.
On the contrary, we propose to make further knowledge distillation with an assistant model $A$, which is able to distill the knowledge for a second time and hence help transfer the information from $T$ to $S$ more sufficiently.
A very recent work, called TAKD \cite{mirzadeh2019improved}, proposed to improve KD method by introducing intermediate teacher assistant model ($TA$).
Specially, TAKD transfers knowledge from $T$ to $S$ with two steps (\emph{i.e.}, first from $T$ to $TA$, then from $TA$ to $S$), which are independent from each other.
Differently, however, RKD employs the assistant model $A$ to learn the residual error between $S$ and $T$, such that $A$ can acquire information from both of them.


\textbf{Residual Learning.}
Residual learning is an effective learning scheme, which is firstly adopted to CNN by \citet{he2016deep}.
After that, this strategy has been applied to various tasks, such as visual question-answering \cite{kim2016multimodal}, stereo matching \cite{pang2017cascade}, image denoising \cite{zhang2017beyond}, and image super resolution \cite{kim2016accurate,wang2018esrgan}.
In general, the core thought is based on the hypothesis that learning the residual is easier than optimizing the target function directly without any reference.
This is consistent with the coarse-to-fine idea, where a problem is solved with a coarser (identity branch) and a finer (residual branch).
This work introduces this principle into KD task.
Unlike prior work that only optimized $S$ (coarser) to learn from $T$, the proposed RKD employs $A$ (finer) to learn the residual error between $S$ and $T$.
In this way, $A$ is able to refine the feature map of $S$, such that the knowledge of $T$ is distilled more completely.
To the best of our knowledge, this is the first work that utilizes residual learning on model compression.

\section{Residual Knowledge Distillation}\label{sec:residual-knowledge-distillation}
\begin{figure}[t]
  \centering
  \includegraphics[width=1.0\linewidth]{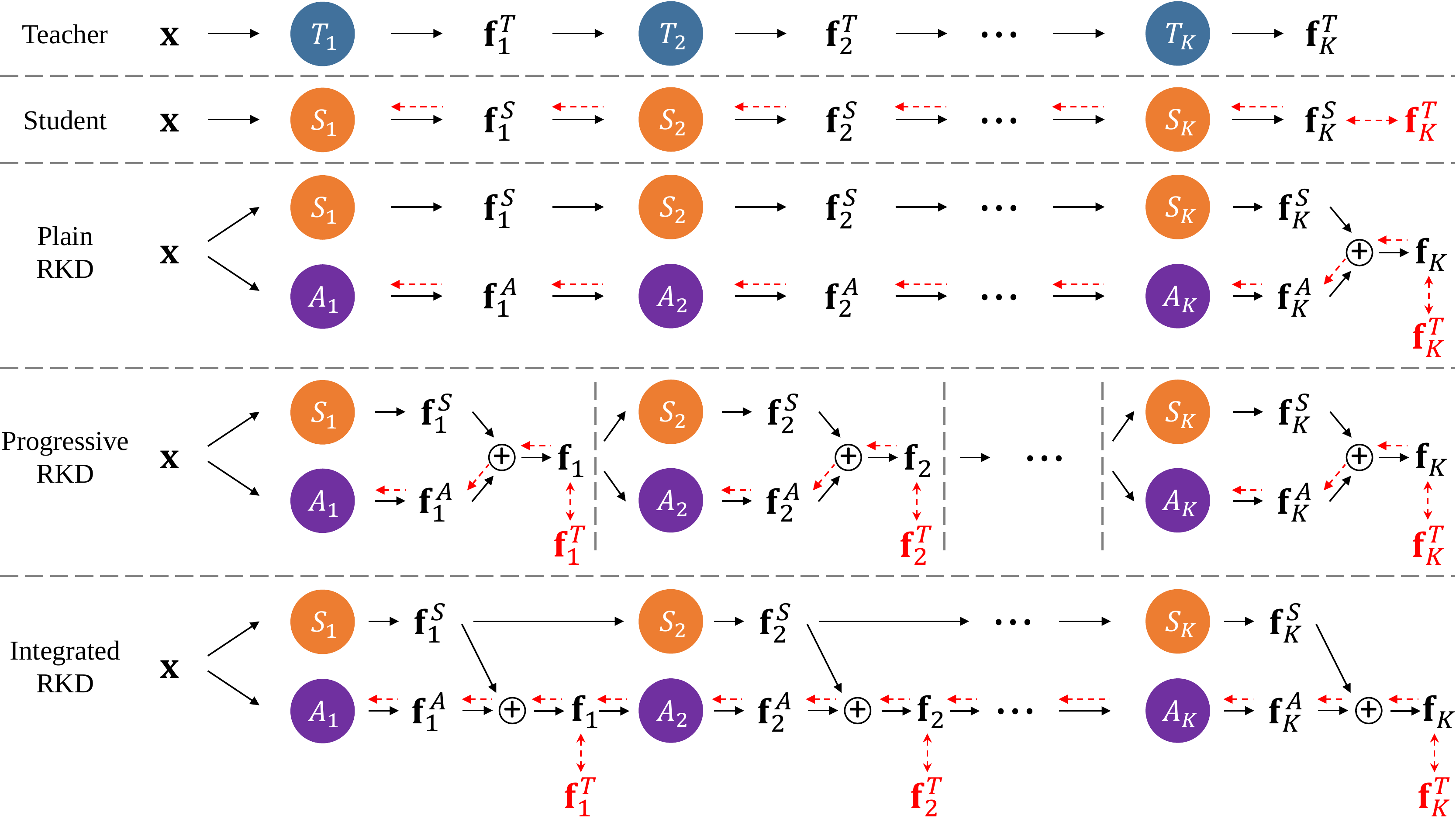}
  \captionsetup{font=small}
  \caption{
    The first two rows illustrate how the knowledge is distilled from teacher ($T$) to student ($S$), both of which are divided into $K$ blocks.
    To achieve more accurate information transfer, we separate the original student model into two \emph{sub-networks}, $S$ and $A$, and introduce ideas of residual learning (Plain RKD) and progressive learning (Progressive RKD) into the above process.
    Similar as $T$ and $S$, $A$ is also divided into $K$ blocks.
    Last row show the integration of these two strategies, resulting in an end-to-end training manner.
    Black arrows and red dashed arrows represent forward and backward propagation respectively, and the red dashed two-way arrows indicate the supervision from $T$.
    Better viewed in color.
  }
  \label{fig:framework}
\end{figure}

We formulate RKD by employing two models, \ie student ($S$) as well as the assistant ($A$), to get knowledge from teacher ($T$) in a complementary manner.
As shown in Fig.\ref{fig:framework}, besides the plain RKD where $A$ only learns the residual error corresponding to the final feature map, we also involve the idea of progressive learning (Progressive RKD) and then integrate it into an end-to-end training scheme (Integrated RKD).
In addition, we explore an efficient way to derive the network structures of $S$ and $A$ from a pre-designed model, such that the total computational cost will not increase after introducing $A$.
More details will be discussed in the following sections.

\subsection{Knowledge Distillation}\label{subsec:knowledge-distillation-method}
Knowledge distillation (KD) method typically employs a student $S(\cdot)$ to learn from a well-trained teacher model $T(\cdot)$, aiming at reproducing the predictive capability of $T$.
In other words, given an image-label pair $(\x, y)$, $T$ will make a prediction $\hat{y}^T = T(\x)$, and $S$ is trained with the purpose of outputting similar result as $\hat{y}^T$.
Here, the prediction made by $S$ is denoted as $\hat{y}^S = S(\x)$.

To achieve this goal, KD targets at exploring a way to extract the information contained in a CNN model and then push the information of $S$ to be as close to that of $T$ as possible.
Accordingly, KD can be formulated as
\begin{align}
  \min_{\Theta_S}\loss_S = d(\psi(T(\cdot), \Theta_T), \psi(S(\cdot), \Theta_S)),  \label{eq:KD}
\end{align}
where $\Theta_T$ and $\Theta_S$ are the trainable parameters of $T$ and $S$ respectively.
$\psi(\cdot, \cdot)$ is the function that helps define the knowledge of a particular model, and $d(\cdot, \cdot)$ is the metric to measure the distance between the knowledge of two models.

Note that, only $\Theta_S$ in Eq.\eqref{eq:KD} is updated, since $T$ is assumed to have already been optimized with ground-truth data.
For classification tasks, cross-entropy loss is used as the objective function
\begin{align}
  \min_{\Theta_T}\loss_T = -\sum_{i=1}^{N}y_i\log\hat{y}_i^T,  \label{eq:task}
\end{align}
where $N$ is the number of categories,
$y$ is an $N$-dimensional one-hot vector indicating the ground-truth label, while $\hat{y}^T$ is the soft probabilities predicted by $T$.

\subsection{Residual Learning with Assistant (Plain RKD)}\label{subsec:residual-learning-with-assistant}
In this work, we treat feature maps of a CNN model as the underlying knowledge.
Generally, a model can be divided into a set of blocks, and the output of each block is considered as a hidden feature map.
Taking teacher in Fig.\ref{fig:framework} as an example, $T$ consists of $K$ blocks, $\{T_i(\cdot)\}_{i=1}^K$, and possesses $K$ hidden feature maps, $\{\f_i^T\}_{i=1}^K$, correspondingly.
Besides the $K$ blocks shown in Fig.\ref{fig:framework}, $T$ also employs a classifier, \ie a fully-connected layer activated by softmax function, to convert the final feature map $\f_K^T$ to soft label prediction $\hat{y}^T$.
However, inspired by \citet{gao2018embarrassingly}, classifiers of $T$ and $S$ share the same structure as well as equal learning capacity, and hence they are excluded from the knowledge distillation process.
%
%
In this way, Eq.\eqref{eq:KD} can be simplified as
\begin{align}
  \min_{\Theta_S}\loss_S = ||\f_K^T - \f_K^S||_2^2,  \label{eq:student}
\end{align}
where $||\cdot||_2$ denotes the $l_2$ distance between two tensors.

In other words, $S$ attempts to produce identical feature map as $T$ such that they can achieve comparable performance.
However, considering the substantial gap between the representation capacities of $S$ and $T$, it is not easy to fit sufficiently well the underlying knowledge captured by the feature maps of $T$ with just $S$ alone.
To solve this problem, we employ an assistant model $A$, to aid $S$ in this mimicking process, which is shown in Fig.\ref{fig:framework} (Plain RKD).
Specifically, $A$, also with $K$ blocks, takes the image $\x$ as input and is optimized to learn the residual error between $\f_K^S$ and $\f_K^T$ with
\begin{align}
  \min_{\Theta_A}\loss_A = ||(\f_K^T - \f_K^S) - \f_K^A||_2^2.  \label{eq:ta}
\end{align}
Then, after $A$ and $S$ reaching their optima respectively, the feature map summed up with the residual error, $\f_K = \f_K^S + \f_K^A$, will be finally used for inference.

By introducing $A$, the knowledge is distilled via two phases, where $S$ is firstly optimized with Eq.\eqref{eq:student} to mimic the hidden feature map of $T$, and then the parameters of $A$ are updated with Eq.\eqref{eq:ta} at the residual learning stage to refine the feature map learned by $S$.
Note that \emph{only} one model, either $S$ or $A$, is trained at each phase.
Although trained separately, $A$ share the same goal with $S$, which is to approximate $\f_K^T$ with $\f_K$.
Consequently, $A$ complements with $S$ to improve the performance by picking up the information which is missed in the first phase.

\subsection{Progressive Learning (Progressive RKD)}\label{subsec:progressive-learning}
As mentioned above, there are a total of $K$ hidden feature maps in each model, however, only the final one is used as reference in Plain RKD.
In this way, even with the help of $A$, some low-level information may still get lost after distillation.
To inherit knowledge from $T$ more completely, we propose Progressive RKD by introducing the idea of progressive learning \cite{wang2018progressive,gao2018embarrassingly}, as shown in Fig.\ref{fig:framework}.

Intuitively, we perform RKD within each block to facilitate the knowledge transfer process.
More concretely, $S$ and $A$ are trained block by block, while the training procedure of each block is same as that in Sec.\ref{subsec:residual-learning-with-assistant}.
For example, $S_1$ is first trained to mimic $\f_I^T$, then $A_1$ learns the residual error \wrt the first intermediate feature map.
After that, the residual-error-involved feature map $\f_1$ is treated as input to the second block for following training.
With this strategy, $S$, in addition with $A$, is able to get both high-level and low-level information from $T$, resulting in better mimicking.

\subsection{Integrating Two Strategies (Integrated RKD)}\label{subsec:integrating-two-strategies}
%
%
We further propose another variant of RKD, called Integrated RKD, by fusing the aforementioned residual learning and progressive learning strategies, as shown in Fig.\ref{fig:framework}.
There are two appealing advantages.
Firstly, compared to Plain RKD in Sec.\ref{subsec:residual-learning-with-assistant}, $A$ is able to capture the information that $S$ fail to learn from multiple levels.
Secondly, it significantly simplifies the training procedure in Sec.\ref{subsec:progressive-learning} for two reasons:
i) the input feature map of each block in $S$ is no longer refined by the residual error produced by $A$, which enables the end-to-end pre-training of $S$ with Eq.\eqref{eq:student};
%
%
%
%
ii) $A$ obtains multi-level supervision from $T$ simultaneously such that all blocks of $A$ can also be optimized at one time with
%
%
\begin{align}
  \min_{\Theta_A}\loss_A = \sum_{i=1}^{K}||(\f_i^T - \f_i^S) - \f_i^{A}||_2^2.  \label{eq:new-a}
\end{align}

\subsection{Model Separation}\label{subsec:model-separation}
RKD is a simple yet novel model compression method that is able to further distill the knowledge from $T$ with the help of $A$.
Accordingly, a crux of RKD is how to choose the model structure of $A$.
One feasible solution is to employ an existing model to serve as $A$, but the computation will increase because of the introduction of additional model.
To solve this problem, we propose a method to derive $S$ and $A$ from a given model while keeping the total computational cost maintained.

The basic idea is to divide a wide model into two thinner ones by reducing the number of channels, while both derived models share the same model structure, \eg number of layers and kernel size of each layer, as the original one.
More concretely, for a particular layer of the source model, we allocate different number of convolutional kernels (channels) to $S$ and $A$ respectively.
However, to better control the computational cost, we use floating-point operations (FLOPs) as measurement instead of the total number of kernels.
Taking four-to-one proportion as an instance, a layer with 100 kernels in the source model is not always separated to two layers with 80 and 20 kernels, because the inputs of these two layers also have different number of channels.
Instead, we compute how many kernels should be used in each target model (\emph{e.g.} $\sqrt{0.8}\times100\approx89$ and $\sqrt{0.2}\times100\approx45$) based on the FLOPs required in the current layer.
In this way, the total model capacities before and after the separation are guaranteed to be almost equal.
Then, the separation is made layer by layer, resulting in two sub-models at last.

\subsection{Implementation Details}\label{subsec:implementation-details}
In this work, we choose networks from ResNet family \cite{he2016deep} to serve as $T$, $S$, and $A$.
To compute residual error, it requires the feature maps of these models to have the same shape, especially along the channel dimension.
Therefore, an additional convolutional layer with $1\times1$ kernel size is applied to deal with the shape-mismatch case.
According to the structure of ResNet, four sets of residual blocks are employed for feature extraction.
We use the outputs of these blocks as hidden feature maps, whose spatial resolutions are $56\times56$, $28\times28$, $14\times14$, $7\times7$ respectively.
$T$ is trained from scratch with Eq.\eqref{eq:task} using stochastic gradient descent (SGD) optimizer with momentum equal to 0.9.
The learning rate is initialized as 0.1 and drops to 10\% every 30 epoch.
The entire model is trained for 100 epoch.
The same learning policy is applied to both $S$ and $A$.
Recall that when $A$ is optimized to learn the residual error, $S$ is fixed such that all parameters in $A$ and $S$ are only updated once.
%
%

\setlength{\tabcolsep}{2pt}
\begin{table}[t]
  \small\centering
  \captionsetup{font=small}
  \caption{Experiments on ImageNet by separating $A$ from $S$ with different proportions.}
  \label{tab:separation}
  \vspace{2pt}
  \begin{tabular}{ccccc}
    \toprule
    Method & Model & Computational Cost (GFLOPs) & Top-1 (\%) & Top-5 (\%) \\
    \midrule[0.7pt]
    Teacher  & ResNet-34 & 3.4121 & 73.55 & 91.46 \\
    Student  & ResNet-18 & 1.6895 & 69.57 & 89.24 \\
    Baseline & ResNet-18 & 1.6895 & 70.29 & 89.35 \\
    \midrule[0.2pt]
    RKD & ResNet-18-50\% ($S$) \& ResNet-18-50\% ($A$) & 0.8279 + 0.8279 = 1.6558 & 68.51 & 88.53 \\
    RKD & ResNet-18-70\% ($S$) \& ResNet-18-30\% ($A$) & 1.1865 + 0.4985 = 1.6850 & 70.48 & 89.68 \\
    RKD & ResNet-18-80\% ($S$) \& ResNet-18-20\% ($A$) & 1.3490 + 0.3385 = 1.6875 & 71.08 & 89.85 \\
    RKD & ResNet-18-90\% ($S$) \& ResNet-18-10\% ($A$) & 1.5201 + 0.1687 = 1.6888 & \textbf{71.46} & \textbf{90.20} \\
    \midrule[0.7pt]
    Teacher  & ResNet-101 & 7.5230 & 77.99 & 93.87 \\
    Student  & ResNet-50  & 3.8084 & 76.01 & 92.98 \\
    Baseline & ResNet-50 & 3.8084 & 76.63 & 93.02 \\
    \midrule[0.2pt]
    RKD & ResNet-50-90\% ($S$) \& ResNet-50-10\% ($A$) & 3.4270 + 0.3815 = 3.8085 & \textbf{77.69} & \textbf{93.45} \\
    \bottomrule
  \end{tabular}
\end{table}

\section{Experiments}\label{sec:experiment}
%
%

We first validate the effectiveness of RKD for different models in Sec.\ref{subsec:evaluation-on-model-separation} with the model separation method presented in Sec.\ref{subsec:model-separation}.
Then we make ablation study in Sec.\ref{subsec:ablation-study} to evaluate the three variants of RKD proposed in Sec.\ref{subsec:residual-learning-with-assistant}, Sec\ref{subsec:progressive-learning}, and Sec.\ref{subsec:integrating-two-strategies} respectively, as well as how the capacity of the assistant model $A$ will affect the learning of residual error.
Sec.\ref{subsec:evaluation-on-different-datasets} compares RKD with state-of-the-art KD methods on different image classification datasets, including CIFAR-100 and ImageNet.
Before going into details, we briefly introduce the datasets used in this work.

\textbf{Datasets.}
CIFAR-100 \cite{krizhevsky2009learning} is a commonly used object recognition dataset, which has 100 classes containing 600 images each.
Among them, 500 images are considered as training samples while 100 as testing.
Following prior work \cite{ba2014deep,zagoruyko2017paying,yim2017gift,lee2018self}, we use CIFAR-100 to evaluate the performance of knowledge transfer.
ImageNet \cite{deng2009imagenet} is a 1,000-categories dataset consisting of 1.2M training samples and 50K validation samples.
It is widely applied in image classification task.
This work uses ImageNet to verify whether RKD works well in large-scale experiments.

\subsection{Effectiveness of Model Separation}\label{subsec:evaluation-on-model-separation}
We would like to first validate the effectiveness of RKD in knowledge transfer \emph{without increasing the total computational cost}, as well as find out the best option of the dissection ratio for model separation.
Here, Integrated RKD method is applied.
Together with a baseline model where student learns to mimic feature maps of teacher directly without the help of assistant, we train several models on ImageNet with different separation proportions and compare their performances and computational costs in Tab.\ref{tab:separation}.
%
%
%
%
%
From Tab.\ref{tab:separation}, we obtain four conclusions as following.
i) After separation, the two derived models (\emph{i.e.}, $S$ and $A$) always have nearly the same computational cost as the original model regardless of the separation ratio.
%
%
ii) RKD works well when $S$ dominates the distillation process, and we empirically found that 90\%-10\% is the best choice.
%
%
In other words, $S$ should learn the majority knowledge from $T$, while $A$ only complements $S$ by learning missed one.
iii) With \emph{same} model size, RKD is capable of improving the performance of baseline model remarkably (from 70.29\% to 71.46\%).
%
%
iv) The proposed separation method is generally applicable to deeper models, such as ResNet-50.
%
%
%
In the experiment of using ResNet-50 to mimic ResNet-101, RKD with 90\%-10\% separation ratio achieves around 1.0\% improvement over the baseline model, which is even competitive with the teacher model (77.69\% \vs 77.99\%).

\setlength{\tabcolsep}{2pt}
\begin{table}[t]
  \small\centering
  \captionsetup{font=small}
  \caption{
    Ablation study on ImageNet.
    (a) shows the baseline results, (b) compares the performances of the three variants of RKD, while (c) analyzes how RKD is affected by model capacity of assistant model $A$.
    To make better evaluation, all student model are fixed to be the same and an additional model is employed to serve as $A$.
  }
  \label{tab:ablation-study}
  \vspace{2pt}
  \begin{tabular}{cccccc}
    \toprule
    & Method & Model & Top-1 (\%) & Top-5 (\%) & $l_2$ Distance \\
    \midrule
    \multirow{3}{*}{(a)}
    & Teacher & ResNet-34 & 73.55 & 91.46 & - \\
    & Student & ResNet-18 & 69.57 & 89.24 & 2514 \\
    & Baseline \wo $A$ & ResNet-18 & 70.29 & 89.35 &  1517 \\
    \midrule
    \multirow{3}{*}{(b)}
    & Plain RKD & ResNet-18 ($S$) \& ResNet-18-1/2 ($A$) & 71.23 & 89.93 &  881 \\
    & Progressive RKD & ResNet-18 ($S$) \& ResNet-18-1/2 ($A$) & \textbf{71.79} & \textbf{90.25} &  \textbf{693} \\
    & Integrated RKD & ResNet-18 ($S$) \& ResNet-18-1/2 ($A$) & 71.63 & 90.23 &  701 \\
    \midrule
    \multirow{4}{*}{(c)}
    & Plain RKD & ResNet-18 ($S$) \& ResNet-18-1/4 ($A$) & 71.16 & 89.92 &  1034 \\
    & Plain RKD & ResNet-18 ($S$) \& ResNet-18-1/2 ($A$) & 71.23 & 89.94 &   881 \\
    & Plain RKD & \multicolumn{1}{l}{ResNet-18 ($S$) \& ResNet-18 ($A$)} & 71.35 & 90.00 & 839 \\
    & Plain RKD & \multicolumn{1}{l}{ResNet-18 ($S$) \& ResNet-34 ($A$)} & 71.38 & 90.01 & 721 \\
    \bottomrule
  \end{tabular}
\end{table}

\subsection{Ablation Study}\label{subsec:ablation-study}
In this part, we make two ablation studies to i) compare the three variants of RKD from performance and training efficiency aspects, and ii) explore how the assistant model helps learn the residual error.
Basically, we conduct experiments on ImageNet dataset and use ResNet-34 and ResNet-18 as teacher and student respectively.
To better evaluate the role $A$ has played in RKD, we do not separate $A$ from $S$ but simply introduce an additional model.
In this way, all student models are fixed to be the same.
Note that this is \emph{not} the final version of RKD, in which the computational cost will not increase.
%
%
Tab.\ref{tab:ablation-study} shows the results.
Among them, ResNet-18-1/4 indicates a model with same structure as ResNet-18 but only having 1/4 numbers of channels for each layer, and similar is ResNet-18-1/2.
%
%
%
Besides the classification accuracy, the $l_2$ distance between the feature maps of teacher and student (or summed up with that of assistant if applicable) is also reported in Tab.\ref{tab:ablation-study}.
This is the objective function for optimization and hence reflects the knowledge transferring performance to some extent.

\textbf{Different Variants of RKD.}
In this work, we introduce the ideas of residual learning and progressive learning into the knowledge distillation process and come up with a unified structure to fuse these two strategies together, resulting in Plain RKD, Progressive RKD, and Integrated RKD respectively.
We train three independent models on ImageNet dataset with the above transferring strategies and compare their performances in Tab.\ref{tab:ablation-study} (b).
We can easily tell that both Progressive RKD and Integrated RKD surpass Plain RKD with more than 0.4\% accuracy, suggesting that RKD benefits a lot from the low-level supervision from $T$.
The $l_2$ distance between feature maps also lead to the same conclusion.
%
%
Although Progressive RKD achieves best result, training with it is tedious, since blocks in $S$ and $A$ are required to be updated alternatively.
On the contrary, Integrated RKD is able to achieve similar result but with much simpler training procedure.
Accordingly, in the following parts, RKD refers to Integrated RKD if not specified.

\textbf{Capacity of Assistant Model.}
The rational behind RKD is to use an assistant model to help capture the missing information by learning the residual error between the feature maps of $S$ and $T$.
In this part, we would like to verify that such residual knowledge is actually learnable, and further explore how much effort (computations) should be put to learn such knowledge.
For this purpose, we use models with various capacities to serve as $A$ based on a same \emph{fixed} student model, and Plain RKD is applied.
Tab.\ref{tab:ablation-study} illustrates the comparison results, from which we have three observations.
i) All four experiments perform better than the baseline in Tab.\ref{tab:ablation-study} (a), suggesting that $A$ indeed complements $S$ in learning the missing knowledge. 
The last column in Tab.\ref{tab:ablation-study} (a) also tells that by learning the residual error, $A$ successfully minimizes the $l_2$ distance between the feature maps of $S$ and $T$.
%
%
%
ii) RKD is able to achieve better accuracy with negligible computational cost increments.
For example, we get 0.87\% higher accuracy by employing ResNet-18-1/4 as $A$, which only consumes 6.25\% computing resources compared to $S$.
This is because learning residual error is more easily \cite{he2016deep}.
This conclusion is consistent with the observation in Sec.\ref{subsec:evaluation-on-model-separation}.
iii) Increasing the model size of $A$ will not always result in performance gain (the last two rows in Fig.\ref{tab:ablation-study} (c)).
Interestingly, even adopting ResNet-34 (same as teacher) as $A$, the accuracy is still far behind that of $T$.
Therefore, $A$ does not require tremendous computing resources, yet a lightweight model is substantial for $A$ to mimic the residual error.

%
%
%
%
%
%
%

\setlength{\tabcolsep}{3pt}
\begin{table}[t]
  \small\centering
  \captionsetup{font=small}
  \caption{Comparison results of image classification on CIFAR-100 and ImageNet datasets.}
  \label{tab:comparison}
  \vspace{2pt}
  \begin{tabular}{ccccccc}
    \toprule
    \multirow{2}{*}{Method} & \multirow{2}{*}{Model} & \multirow{2}{*}{Computational Cost (GFLOPs)} & \multicolumn{2}{c}{CIFAR-100} & \multicolumn{2}{c}{ImageNet} \\ \cmidrule(lr){4-5}\cmidrule(lr){6-7}
    & & & Top-1 (\%) & Top-5 (\%) & Top-1 (\%) & Top-5 (\%) \\
    \midrule
    Teacher & ResNet-34 & 3.4121 & 73.05 & 91.55 & 73.55 & 91.46 \\
    Student & ResNet-18 & 1.6895 & 68.06 & 89.60 & 69.57 & 89.24 \\
    \midrule
    KD      & ResNet-18 & 1.6895 & 72.39 & 91.06 & 70.76 & 89.81 \\
    FitNets & ResNet-18 & 1.6895 & 71.66 & 90.28 & 70.66 & 89.23 \\
    AT      & ResNet-18 & 1.6895 & 70.74 & 90.04 & 70.73 & 90.04 \\
    \midrule
    RKD     & ResNet-18 & 1.6888 & 72.82 & 91.41 & 71.46 & 90.20 \\
    RKD + AT & ResNet-18 & 1.6888 & \textbf{72.96} & \textbf{91.44} & \textbf{71.54} & \textbf{90.26} \\
    \bottomrule
  \end{tabular}
\end{table}

\begin{figure}[t]
  \centering
  \includegraphics[width=1.0\linewidth]{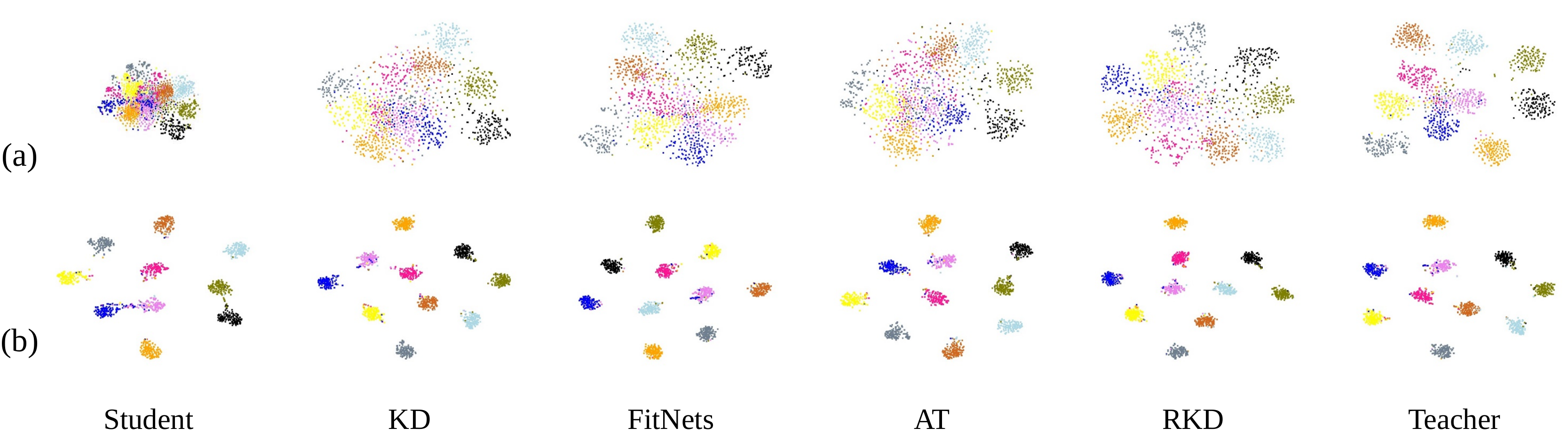}
  \captionsetup{font=small}
  \caption{
    Visualization of feature maps on CIFAR-100 dataset.
    Different columns show the outputs from different KD methods.
    (a) indicates the hidden feature map produced by the third block ($14\times14$ resolution), while (b) comes from the fourth block ($7\times7$ resolution).
    Better viewed in color.
  }
  \label{fig:feature-map}
\end{figure}

\subsection{Evaluation on Different Datasets}\label{subsec:evaluation-on-different-datasets}
This section compares RKD with state-of-the-art knowledge transfer methods, including KD \cite{hinton2014distilling}, FitNets \cite{romero2015fitnets}, and AT \cite{zagoruyko2017paying}, on two widely used datasets, CIFAR-100 and ImageNet.
%
%
Here, to make fair comparisons, we set $T=4$ and $\lambda=16$ for KD method following \cite{hinton2014distilling}.
FitNets uses the last layer of the second residual block in ResNet as hint layer, and AT uses four hidden feature maps (\emph{i.e.}, similar as our RKD) from each residual block to compute attention maps.
We also perform RKD together with AT to verify how the idea of residual learning can assist other methods.

\textbf{Evaluation on CIFAR-100.}
Like existing work \cite{romero2015fitnets,zagoruyko2017paying} did, we firstly carry out small-scale experiments on CIFAR-100 dataset.
Tab.\ref{tab:comparison} shows the comparison results, where RKD outperforms the baseline model with 2.75\% higher accuracy and nearly achieves same performance as teacher model.
We also beat the second competitor by 0.42\% accuracy.
In addition, we randomly pick 10 classes and visualize the corresponding hidden feature maps produced by different models using t-SNE \cite{maaten2008visualizing}.
Fig.\ref{fig:feature-map} shows the feature maps from the third block and the fourth block (final feature map).
We can tell that feature maps from RKD, especially from the third block (Fig.\ref{fig:feature-map} (a)), are more discriminative than other methods, benefiting from the residual error learned by $A$.
In other words, $A$ refines the feature maps outputted by $S$ so as to enhance the predictive capability of student model.

\textbf{Evaluation on ImageNet.}
We also conduct larger-scale experiments on ImageNet dataset, whose results are shown in Tab.\ref{tab:comparison}.
We can easily conclude that RKD improves the baseline model with 1.9\% accuracy, significantly surpassing the alternative KD methods.
Furthermore, by cross-comparing the results on CIFAR-100 and ImageNet, we find out that other methods may perform inconsistently on different datasets.
For example, AT works well on ImageNet but fails on CIFAR-100.
By contrast, RKD shows stronger stability and robustness.

We also distill the knowledge from teacher by combining RKD with AT \cite{zagoruyko2017paying}.
Specifically, $S$ aims at mimicking the attention map, instead of feature map, from $T$, and $A$ is still optimized to learn the residual error.
Last row in Tab.\ref{tab:comparison} suggests that the idea of RKD can be also generalized to other KD methods as long as the residual error is well defined.
To this end, we believe RKD is a promising solution to model compression problem that is worth exploring.

\section{Conclusion}\label{sec:conclusion}
This paper presents a novel model compression approach, called RKD, which further distills the knowledge from teacher model with an assistant ($A$).
By mimicking the residual error between the feature maps of $S$ and $T$, $A$ is able to complement $S$ with the missing information and thus improves the performance significantly.
In addition, we also find a way to split a model apart to $S$ and $A$, such that the computational cost will not increase after introducing $A$.
Numerous experimental results demonstrate the effectiveness and efficiency of RKD.

{\small
\bibliographystyle{abbrvnat}
\bibliography{ref}
}

\end{document}